# RiskNet: Interaction-Aware Risk Forecasting for Autonomous Driving in Long-Tail Scenarios

Qichao Liu[a,b], Heye Huang[b]*, Shiyue Zhao[c], Lei Shi[b], Soyoung Ahn[b], Xiaopeng Li[b]

*[a] School of Transportation, Southeast University, 2 Si Pai Lou, Nanjing, 210096, China*
*[b] Department of Civil & Environmental Engineering, University of Wisconsin-Madison, Madison, 53706, USA*
*[c] School of Vehicle and Mobility, Tsinghua University, Beijing, 100084, China*

**Abstract**

Ensuring the safety of autonomous vehicles (AVs) in long-tail scenarios remains a critical challenge, particularly under high uncertainty and complex multi-agent interactions. To address this, we propose RiskNet, an interaction-aware risk forecasting framework, which integrates deterministic risk modeling with probabilistic behavior prediction for comprehensive risk assessment. At its core, RiskNet employs a field-theoretic model that captures interactions among ego vehicle, surrounding agents, and infrastructure via interaction fields and force. This model supports multidimensional risk evaluation across diverse scenarios (highways, intersections, and roundabouts), and shows robustness under high-risk and long-tail settings. To capture the behavioral uncertainty, we incorporate a graph neural network (GNN)-based trajectory prediction module, which learns multi-modal future motion distributions. Coupled with the deterministic risk field, it enables dynamic, probabilistic risk inference across time, enabling proactive safety assessment under uncertainty. Evaluations on the highD, inD, and rounD datasets, spanning lane changes, turns, and complex merges, demonstrate that our method significantly outperforms traditional approaches (e.g., TTC, THW, RSS, NC Field) in terms of accuracy, responsiveness, and directional sensitivity, while maintaining strong generalization across scenarios. This framework supports real-time, scenario-adaptive risk forecasting and demonstrates strong generalization across uncertain driving environments. It offers a unified foundation for safety-critical decision-making in long-tail scenarios.

Keywords: Risk forecasting; Interaction field; Graph neural networks; Uncertainty modeling

## 1. Introduction

*1.1 Motivation*

Autonomous Driving Systems (ADS) are emerging as a foundational component of intelligent transportation systems, aiming to improve traffic efficiency, reduce human errors, and optimize the allocation of road resources (Guo et al., 2020; Wang et al., 2021). Among its core modules, risk perception and quantification are essential for ensuring the safety and reliability of ADS under real-world traffic conditions. Real-world environments involve highly uncertain behaviors, heterogeneous participants, and non-structured road features (Cao et al., 2022; Yao et al., 2024). To further investigate the structural characteristics of traffic risk, we conducted a statistical analysis based on the Fatality Analysis Reporting System (FARS), a nationwide database published by the U.S. National Highway Traffic Safety Administration (NHTSA) ("Fatality Analysis Reporting System (FARS) | NHTSA," n.d.). This dataset systematically records heterogeneous sources of risk, including roadway geometry, lane count and separation status, vehicle types, collision modes, and driver behaviors, making it a representative foundation for analyzing multi-source interactions.

---

\* Corresponding author.
*E-mail address: hhuang468@wisc.edu*. (H.Huang)

**Fig. 1** presents the joint distributions of key risk factors. As shown in **Fig. 1(a)**, fatal crashes are most frequently observed on two-lane undivided roads (4924 cases) and three-lane roads (3799 cases), indicating that urban, low-class, and structurally unprotected roads are particularly accident-prone. In **Fig. 1(b)**, the highest concentration of fatal crashes occurs in frontal collisions involving passenger cars and light trucks, both exceeding 7000 cases, highlighting this combination as the dominant high-risk crash mode.

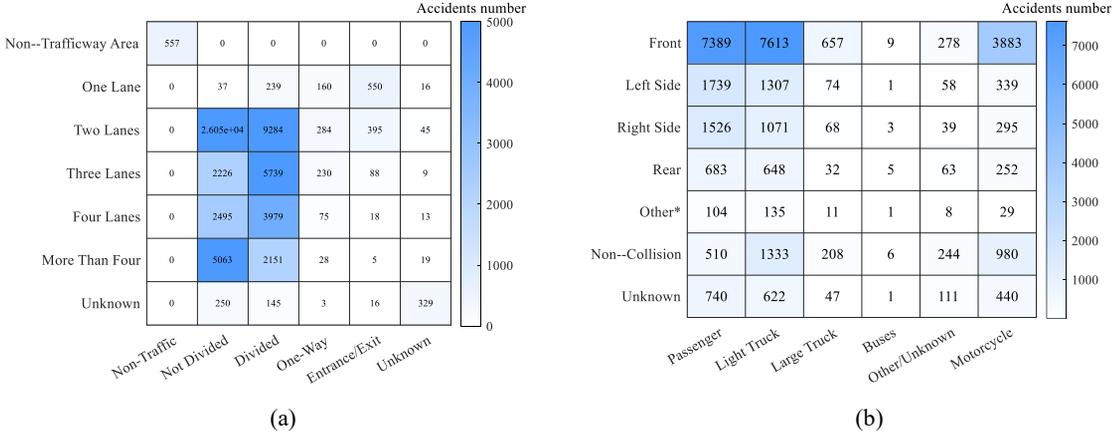

(a) (b)

Fig. 1. Joint distribution analysis (a) Distribution of fatal crashes across lane count and road separation type; (b) Distribution of fatal crashes across collision types and vehicle categories

These observations suggest that crash risk arises from the interaction of multiple heterogeneous sources, including road infrastructure, vehicle characteristics, and collision behaviors (Kolekar et al., 2020; Wang et al., 2020). Moreover, the risk pattern is not unidimensional, it spans across infrastructure configuration × behavioral mode × vehicle type, reflecting a multi-source, multi-dimensional coupling structure. This analysis provides empirical evidence for the necessity of an integrated risk quantification framework capable of capturing such complex interaction dynamics in real-world autonomous driving scenarios.

Building on this analysis, we introduce and define the concept of long-tail scenarios (Zhou et al., 2023), referring to rare but safety-critical driving conditions that occur infrequently in historical data yet pose substantial safety threats when they arise. Such scenarios are typically characterized by high behavioral uncertainty, non-standard road configurations, difficult perception conditions, and complex multi-agent interactions. Examples include high-speed emergency lane changes, pedestrian emergence under occlusion, and vehicle conflicts at busy urban intersections. Despite their low occurrence probability, these scenarios present the most severe challenges to system stability and decision-making reliability (Huang et al., 2025).

Traditional risk assessment metrics, such as Time to Collision (TTC) (Brown, 2005), Time Headway (THW) (Vogel, 2003), and Responsibility-Sensitive Safety (RSS) (Hasuo, 2022), are largely based on deterministic assumptions, and struggle to accommodate the stochasticity and multimodality of real-world traffic behavior. Although deep learning-based trajectory prediction has shown promise (Scheel et al., 2022), a critical gap remains in mapping probabilistic prediction outputs into structured, interpretable, and operational risk measures.

This motivates the development of RiskNet, a unified interaction-aware risk forecasting framework that combines physics-based interaction modeling with data-driven uncertainty prediction, enabling more effective risk perception and evaluation in complex long-tail driving scenarios.

## 1.2 Main Contributions

To address these challenges, this study introduces RiskNet, an interaction-aware, multidimensional risk prediction framework that integrates deterministic assessment with deep learning–based probabilistic prediction.

RiskNet is designed to effectively capture driving risks for autonomous vehicles (AVs) in long-tail scenarios, where uncertainty and complex interactions dominate. The main contributions are as follows:
- We propose RiskNet, a unified framework that integrates physics-inspired interaction field modeling with data-driven trajectory prediction, enabling temporal risk forecasting in complex traffic.
- We develop a field-theoretic risk model that integrates interaction fields, directional force propagation, and anisotropic correction to capture inter-agent dynamics and enable accurate assessment in multi-agent, multi-directional, and long-tail scenarios.
- We evaluate the framework on highD, inD, and rounD datasets involving long-tail scenarios. Results show that our model outperforms TTC, THW, RSS, and NC Field in accuracy, spatial sensitivity, and computational efficiency, with strong generalization to real-world conditions.

*1.3 Paper Organization*

The remainder of this paper is organized as follows. Section 2 reviews related work on risk assessment and trajectory prediction in autonomous driving. Section 3 introduces the overall architecture of the proposed RiskNet framework. Section 4 details the methodology, including the deterministic multisource risk modeling module and the uncertainty-aware trajectory prediction and risk evaluation components. Section 5 reports experimental results on multiple naturalistic driving datasets and compares model performance under both deterministic and uncertain scenarios. Finally, Section 6 concludes the paper and discusses directions for future research.

## 2. Related Works

*2.1 Risk assessment*

The safety of AVs is influenced by multiple sources of risk, including ego-vehicle control uncertainty, interactions with surrounding traffic participants, roadway environmental complexity, and system-level robustness. A range of risk assessment approaches has been proposed in the literature, which can be broadly categorized into kinematics-based, collision probability-based, and artificial potential field (APF)-based methods (Goerlandt and Reniers, 2016; Li et al., 2020).

Kinematic and dynamic model-based methods rely on predefined motion assumptions, such as constant velocity (CV), constant acceleration (CA), or fixed headway, and compute risk indicators based on temporal or spatial logic (Zhou and Zhong, 2020). Common metrics include Time to Collision (TTC) (Köhler, 2024), Time Headway (THW), Time to Steer (TTS) (Wang et al., 2022), and Post Encroachment Time (PET) (Chang and Jazayeri, 2018). These methods are straightforward and physically interpretable, and they perform well in structured environments like highways (Cheng et al., 2025). However, they are often inadequate in unstructured or highly interactive urban settings, where behavioral diversity and long-tail scenarios challenge their generalizability and accuracy.

Collision probability-based methods seek to estimate risk by predicting future trajectory distributions for traffic participants and computing the likelihood of collision with the ego vehicle (Wang et al., 2024). Typically, such methods involve sampling-based prediction, probabilistic density estimation, and trajectory overlap detection (Gindele et al., 2013; Ortiz et al., 2023; Schulz, 2021). While these approaches are theoretically capable of incorporating behavioral and environmental uncertainty, they often suffer from high computational cost and rely heavily on dense and accurate trajectory data, which limits their scalability for real-time applications.

APF-based methods simulate risk as physical force fields, where other vehicles, pedestrians, and obstacles exert repulsive or attractive forces on the ego vehicle. For example, research proposed an electric-field-inspired model that generates avoidance trajectories based on potential intensity and conducted trajectory planning (Bounini et al., 2017; Semsar-Kazerooni et al., 2017). Wang et al. further introduced the concept of a driving safety field, integrating kinetic, potential, and behavioral fields to describe complex human–vehicle–road interactions in a unified framework (Wang et al., 2015). Huang et al. further extended this framework by introducing a probabilistic intention field to model stochastic lane-changing behavior, improving the detection of potential conflicts in mixed-traffic scenarios (Huang et al., 2020). APF-based methods are particularly effective in capturing real-time interaction characteristics and spatial conflict zones. However, they tend to fall into local optima in complex environments and often lack the ability to model stochastic human behaviors and policy-level uncertainty, making them less effective in multi-agent, dense interaction urban scenarios.

In summary, although current methods demonstrate strong performance in structured environments, they often struggle to scale and generalize in the face of uncertainty and complex interactions. Developing a unified, physically grounded, and probabilistically informed risk assessment framework remains a core challenge for real-world autonomous driving applications.

## 2.2 Trajectory prediction

In complex and highly dynamic traffic environments, AVs must not only assess current risks but also anticipate future hazards driven by the behaviors of surrounding road users. Trajectory prediction plays a vital role in this process, aiming to infer the future motions of nearby agents based on their past trajectories, relative dynamics concerning the ego vehicle, and environmental context (Leon and Gavrilescu, 2021; Liu et al., 2019). Broadly, existing trajectory prediction approaches can be categorized into model-based and data-driven methods.

Model-based approaches rely on predefined kinematic or dynamic assumptions, using historical motion states to extrapolate future trajectories (Bahram et al., 2016). Common models include constant velocity (CV), constant acceleration (CA), constant turn rate and velocity (CTRV), and constant turn rate and acceleration (CTRA) (Karle et al., 2022). While computationally efficient and suitable for short-horizon prediction, these methods often struggle with long-term forecasting and fail to capture the behavioral complexity and interaction dynamics prevalent in urban traffic.

In contrast, data-driven methods, particularly those based on deep learning, have shown strong potential in learning complex motion patterns and social interactions from data (Deo and Trivedi, 2018; Gupta et al., 2018). Recurrent architectures, such as LSTMs, are widely adopted for their temporal modeling capabilities (Altché and Fortelle, 2017). For example, an attention-aware social graph transformer network that integrates graph convolution and transformer modules to improve long-term trajectory prediction in mixed scenarios (Y. Liu et al., 2024). However, LSTM-based models tend to rely on fixed, often unimodal distribution assumptions (e.g., Gaussian), which limits their ability to represent multi-modal behaviors in diverse driving scenarios. Meanwhile, Generative Adversarial Networks (GANs) have been employed to produce socially and physically plausible trajectories under complex conditions (Lin et al., 2023). Some attention-based GAN models have improved prediction diversity, but challenges remain in training stability, probabilistic interpretability, and control over generated outputs.

To address these limitations, MTP-GO (graph-based probabilistic multi-agent trajectory prediction with neural ODEs) introduces a principled framework that combines temporal graph neural networks for interaction modeling with neural ordinary differential equations (Neural ODEs) for continuous-time state evolution (Westny et al., 2023). Additionally, the integration of mixture density networks and Kalman filtering enables accurate and interpretable multimodal predictions. Inspired by MTP-GO, the trajectory prediction module in our framework incorporates graph neural networks (GNNs), captures both discrete social dependencies and continuous motion evolution. This architecture enables the generation of multi-modal trajectories with calibrated uncertainty, which are crucial for robust downstream risk assessment, particularly under high-stakes, interactive scenarios.

## 3. RiskNet framework

Fig. 2 illustrates the proposed risk forecasting framework and its application across representative autonomous driving scenarios. The framework integrates deterministic modeling with probabilistic prediction, enabling dynamic risk assessment across diverse scenes, directions, and behavior types.

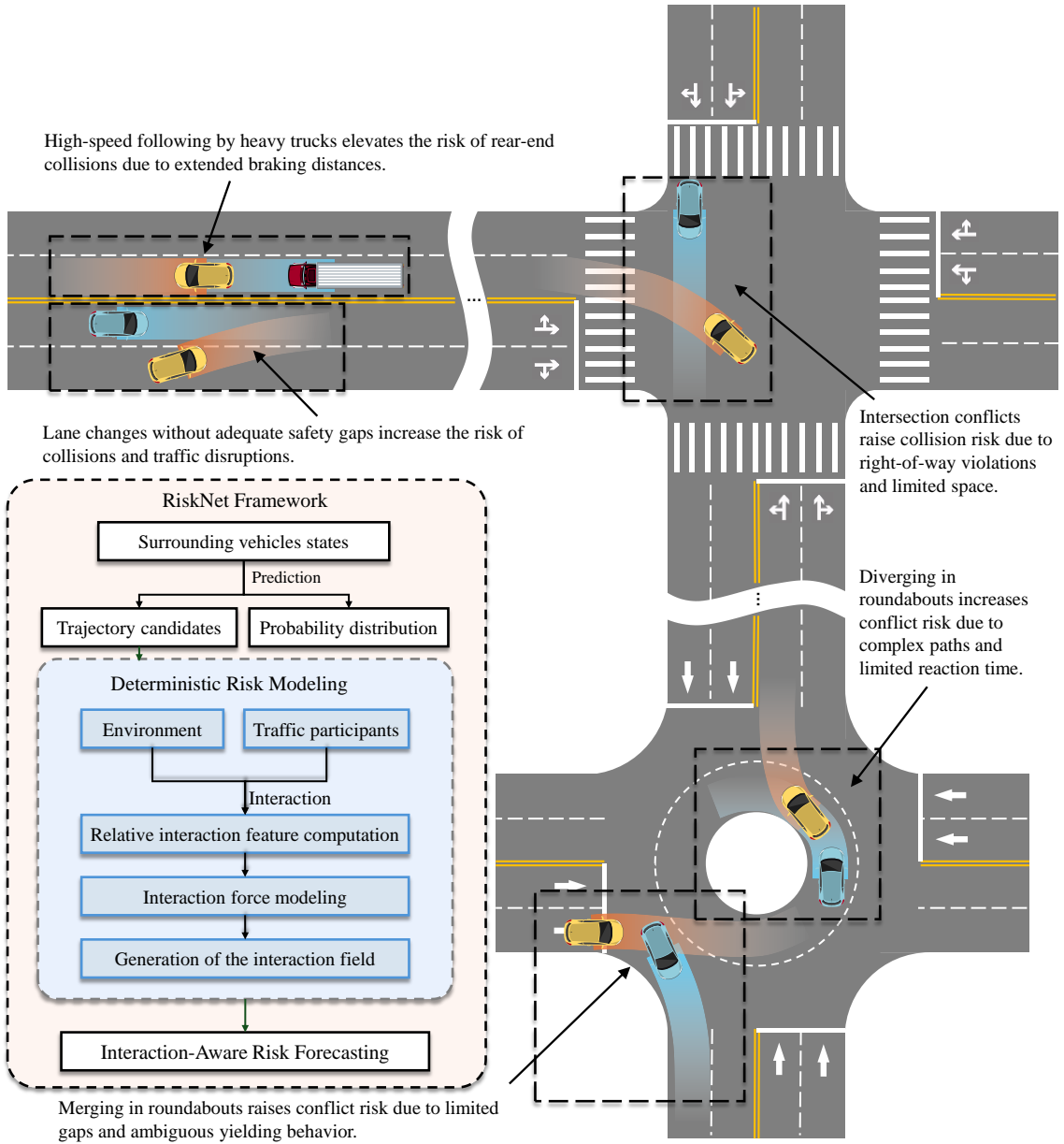

Fig. 2. The proposed RiskNet framework

At its core, the framework incorporates a deterministic multisource risk modeling module, which captures the physical interactions between the ego vehicle, surrounding agents, and the driving environment. Based on an interaction field–force propagation mechanism, this module models directional and agent-specific risk impacts and provides real-time estimation of the driving safety envelope. It identifies potential collision risks across scenarios such as close-following heavy vehicles on highways, insufficient gaps during lane changes, multi-directional conflicts at intersections or roundabouts, and high-interference yielding situations. The model is adaptable and robust to long-tail, high-risk events. Built upon this, a graph neural network (GNN)-based trajectory prediction module captures the uncertainty in surrounding agents' future behaviors, generating multi-modal trajectory distributions. When combined with the deterministic risk field, it enables the identification of high-uncertainty events, such as pedestrian crossings, failed yields, or sudden cut-ins, and simulates their future probabilistic risk evolution. This allows early detection of potential conflict zones caused by uncertain maneuvers (e.g., competitive merges or unsafe yields).

By fusing physics-based modeling with learned behavioral prediction, the framework provides real-time, scenario-adaptive, and generalizable risk assessment, enhancing AV safety in complex and uncertain environments.

## 4. Methodology

### *4.1 Deterministic interaction-aware risk modeling*

In complex traffic environments, driving risk emerges not only from the state of the ego vehicle and its surrounding traffic participants but also from the interplay of roadway conditions, social behaviors, and driver intentions. Consequently, traditional risk assessment approaches—often limited to isolated factors—fall short in accurately modeling the multifaceted realities of real-world traffic scenarios. To address this, we propose a comprehensive multi-source risk assessment framework that captures the coupled dynamics of the human–vehicle–infrastructure system. Grounded in field theory, the framework offers a unified mathematical representation that integrates diverse agent information and characterizes system-wide risk through spatial interaction fields.

#### 4.1.1 Modeling interaction field and force

Traffic risk emerges from interactions among road users and between users and their environment, reflecting inherently systemic relationships. In any traffic scenario involving J participants, we define the set of all traffic agents as:

$$\mathcal{J} = [1, 2, \dots, J] \tag{1}$$

The interaction between the ego vehicle $i$ and any traffic participant $j \in \mathcal{J}$ is represented by an interaction indicator $I_{i,j}$, where $I_{i,j} = 1$ denotes the presence of interaction, and $I_{i,j} = 0$ indicates no interaction.

If such interactions are not properly managed or mitigated, they may lead to collisions. In physics, a collision is characterized as a brief, high-intensity exchange of energy between bodies, and its consequences can be quantified through energy transfer. To evaluate the safety implications of these interactions, we model risk as a field—a continuous spatial representation of potential hazard—where risk is defined as the result of field interactions among traffic entities. This formulation enables us to capture how humans perceive and respond to risk within the traffic environment. Accordingly, the consequences of interaction between any two entities are expressed through an interaction field as:

$$E_i = \frac{1}{2} k_j \cdot C_j \cdot \frac{m_i m_j}{m_i + m_j} \cdot \|\boldsymbol{v}_i(t) - \boldsymbol{v}_j(t)\|^2 \tag{2}$$

where $E_i$ represents the energy transferred by the ego vehicle $i$; $m_i$, $\boldsymbol{v}_i$ denote its mass and velocity vector, respectively, while $m_j$ and $\boldsymbol{v}_j$ correspond to the mass and velocity of the interacting traffic participant $j$. The parameter $k_j \in [0,1]$ characterizes the inherent danger posed by participant $j$, incorporating its physical attributes and type (e.g., pedestrian, cyclist, car, or truck), and reflects its relative contribution to collision risk. The environmental constraint factor $C_j$ accounts for contextual influences—such as speed limits, traffic signals, and lane boundaries—that affect how participant $j$ impacts the ego vehicle.

When interacting with multiple participants, the ego vehicle's total risk is modelled as the cumulative interaction energy:

$$\widetilde{E_i} = \sum_{j=0}^{n} \frac{1}{2} I_{i,j} \cdot k_j \cdot C_j \cdot \frac{m_i m_j}{m_i + m_j} \cdot \|\boldsymbol{v}_i(t) - \boldsymbol{v}_j(t)\|^2 \tag{3}$$

The above formulation shows that each traffic participant contributes to the ego vehicle's risk based on its physical properties, environmental constraints, relative velocity, and interaction intensity—forming a theoretical basis for multi-source risk modeling.

The risk field originates from the ego vehicle's objective to reach its destination safely and efficiently. As it moves, the vehicle is influenced by interaction forces shaped by its position, goal (reflecting driving intent), dynamic state (e.g., speed), and road conditions.

$$F_{ij} = \frac{E_i}{r_{ij}} = \frac{E_i}{\sqrt{\|x_i(t) - x_j(t)\|^2 + \|y_i(t) - y_j(t)\|^2}} \tag{4}$$

where, $F_{ij}$ represents the effective force exerted on the ego vehicle $i$ by traffic participant $j$, serving as a measure of the risk posed by $j$ within the driving environment. Expressed in newtons (N), this force increases with greater relative velocity and shorter distance between the two entities.

When the ego vehicle interacts with multiple participants, the total cumulative force $\widetilde{F_{ij}}$ it experiences is given by:

$$\widetilde{F_{ij}} = \sum_{j=0}^{n} I_{i,j} \frac{E_i}{\sqrt{\|x_i(t) - x_j(t)\|^2 + \|y_i(t) - y_j(t)\|^2}} \tag{5}$$

This force field captures the ego vehicle's exposure to multi-directional, multi-agent influences in complex traffic, allowing the model to adapt to diverse driving scenarios.

4.1.2 Directional risk distribution via the doppler effect

The risk posed by moving objects depends on both distance and the relative velocity vector between the object and the ego vehicle. Interaction outcomes, measured by $E_i$ or $F_{ij}$, are influenced by entities' intrinsic properties (e.g., mass, shape), dynamic states (e.g., speed, acceleration), and environmental factors such as lane markings, signals, and speed limits.

Assume that a traffic participant $j$, located at $(x_j, y_j)$, emits a signal at frequency $f_i$, which is received by the ego vehicle $i$ at position $(x_i, y_i)$ with a perceived frequency $f_i$. A higher $f_i$ indicates a stronger influence of $j$ on the ego vehicle at that location. According to the Doppler effect shown in **Fig.3**, objects approaching in the ego vehicle's direction of motion exert greater perceived risk than those moving away.

$$\frac{f_i}{f_j} = \frac{v_{j,0} \pm v_i(t)}{v_{j,0} \mp v_j(t)} \tag{6}$$

where $v_{j,0}$ is the wave speed, $v_j(t)$ is the velocity of the moving source (participant $j$), $v_i(t)$ is the ego vehicle's velocity. When the ego vehicle moves toward the source, the perceived influence increases (positive sign); when moving away, it decreases (negative sign).

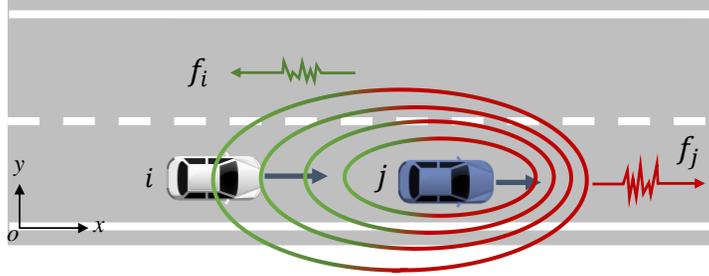

Fig. 3. The doppler effect of a moving object

The Doppler effect illustrates how a moving source alters the frequency of a received signal: the closer the source, the higher the perceived frequency, and vice versa. This property can be leveraged to adjust the directional distribution of the risk field, concentrating risk more heavily in the direction of motion, and aligning better with real-world traffic dynamics. In this formulation, the field originates from the object's center of mass and becomes denser in the forward direction, with a slower gradient decay as proximity increases. Accordingly, we introduce a directional gradient adjustment coefficient, defined as:

$$\alpha_{ij} = \frac{v_{j,0} + v_i(t)\cos\theta_{ij}}{v_{j,0} - v_j(t)\cos\theta_{ij}} \tag{7}$$

The coefficient $\alpha_{i,j}$, derived from the Doppler effect, ensures that the risk field exerts greater influence in the ego vehicle's direction of motion. Here, $\theta_{ij}$ denotes the angle between the velocity vectors of vehicle $i$ and participant $j$. When $\theta_{ij} \to 180°$ (i.e., $j$ is moving in the same direction ahead of $i$), $\alpha_{ij}$ becomes larger, indicating a higher perceived risk from $j$.

To ensure consistency across different scenarios, we define a normalized longitudinal adjustment factor as:

$$\alpha_{ij}^{\text{lon}} = \max(0, \frac{v_{j,0} + v_i(t)\cos\theta_{ij}}{v_{j,0} - v_j(t)\cos\theta_{ij}}) \tag{8}$$

This correction ensures that when $v_j\cos\theta_{ij} > v_{j,0}$, the longitudinal coefficient $\alpha_{ij}^{\text{lon}}$ remains non-negative, avoiding non-physical risk weights.

In addition to longitudinal influence, lateral participants also contribute to risk. Typically, lateral risk is lower than that from forward or rearward vehicles and decreases as the angular deviation increases. We define the lateral risk attenuation factor as:

$$\alpha_{ij}^{\text{lat}} = e^{-\beta \sin^2\theta_{ij}} \tag{9}$$

where, $\beta$ is the lateral decay coefficient, which controls the rate at which lateral risk diminishes, typically ranging from 0.5 to 2 based on empirical observations.

The final directional adjustment coefficient is given by:

$$\alpha_{i,j} = \alpha_{ij}^{\text{lon}} \cdot \alpha_{ij}^{\text{lat}} \tag{10}$$

Considering both longitudinal and lateral directional adjustments, the risk field formulation is updated as:

$$\widetilde{F_{ij}} = \sum_{j=0}^{n} I_{i,j} \cdot \alpha_{ij}^{\text{lon}} \cdot \alpha_{ij}^{\text{lat}} \frac{E_i}{\sqrt{\|x_i(t) - x_j(t)\|^2 + \|y_i(t) - y_j(t)\|^2}} \tag{11}$$

This enhanced expression captures the directional sensitivity of risk, emphasizing forward threats while attenuating lateral ones.

*4.2 Probabilistic interaction-aware risk prediction*

In the previous section, we introduced a fixed-step risk assessment model for evaluating AV safety and triggering collision warnings. However, real-world planning requires longer time horizons, where the complexity and uncertainty of interactive traffic limit the effectiveness of single-step models. To address this, we extend the deterministic model with predictive capabilities, enabling probabilistic risk estimation over future time steps to support trajectory planning.

This extension involves three key challenges: (1) predicting multi-step motions, (2) estimating their probabilities, and (3) assessing risk under uncertainty. To tackle these, we propose an integrated framework that combines probabilistic trajectory modeling, and deterministic risk assessment. Specifically, we develop a graph neural network (GNN)-based motion predictor to infer the uncertain intentions of surrounding agents and output probabilistic future trajectories. The GNN model captures agents' behavioral patterns and outputs trajectory distributions, which are then combined with the deterministic risk field to produce a probabilistic risk map for dynamic risk-aware planning.

4.2.1 Probabilistic behavior evolution

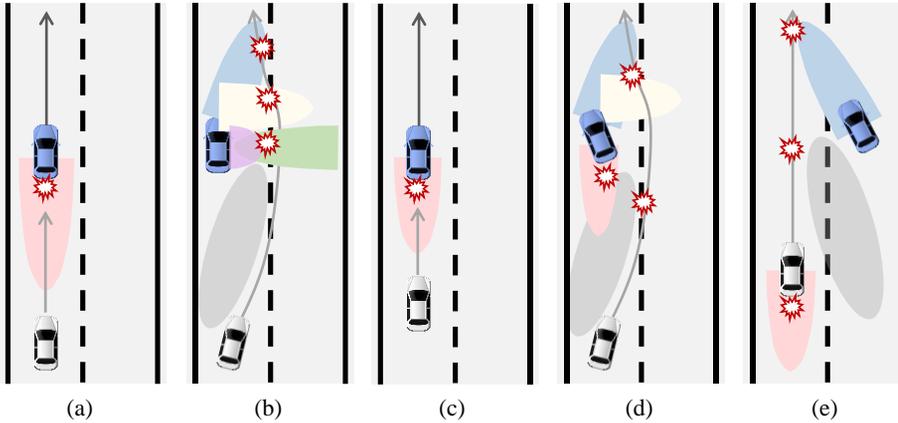

Fig. 4. Predicted interaction risk between ego (white) and surrounding vehicles (blue)

The dynamic and uncertain behavior of traffic participants, along with their interaction with AVs during planning and decision-making, plays a critical role in risk assessment. **Fig. 4** illustrates typical risk zones in diverse scenarios: red indicates rear-end collision risk, yellow denotes lateral interference from adjacent lanes, light blue marks sudden maneuvers, purple represents door-opening hazards, gray shows bicycle avoidance, and green highlights pedestrian crossing risks. Red explosion icons mark potential collision points, and gray lines trace vehicle trajectories. The figure underscores the importance of real-time, multi-source risk perception for safe and adaptive AV decision-making.

4.2.2 Trajectory prediction

To effectively model the future behaviors of traffic participants and facilitate the analysis of uncertainty propagation in driving risk, this study develops a multimodal trajectory prediction module inspired by the design principles of the MTP-GO model. The proposed module integrates Graph Neural Networks (GNNs) with Neural Ordinary Differential Equations (Neural ODEs), enabling the generation of multiple probabilistic future trajectories along with corresponding uncertainty estimates. These outputs serve as essential inputs for subsequent risk assessment tasks. A schematic illustration of the deep learning-based trajectory prediction architecture is shown in **Fig. 5**.

The trajectory prediction problem is formulated as a conditional probabilistic modeling task. Let $\mathcal{H}$ denote the observed history at time $t$, comprising the state sequences of multiple traffic participants over past time steps. The objective is to predict the spatial distribution of each participant over a future horizon of $t_f$ seconds. Formally, the goal is to learn the following conditional distribution:

$$p\left(\left\{(x_j(t+1), \ldots, x_j(t+t_f))\right\}_{j \in \mathcal{N}_i} \middle| \mathcal{H}\right) \quad (12)$$

where, $x_j(t) \in \mathbb{R}^2$ represent the 2D position of the $j$-th traffic participant at time $t$. The set $\mathcal{N}_i$ denotes all surrounding vehicles present in the scene at the prediction time, excluding the ego vehicle.

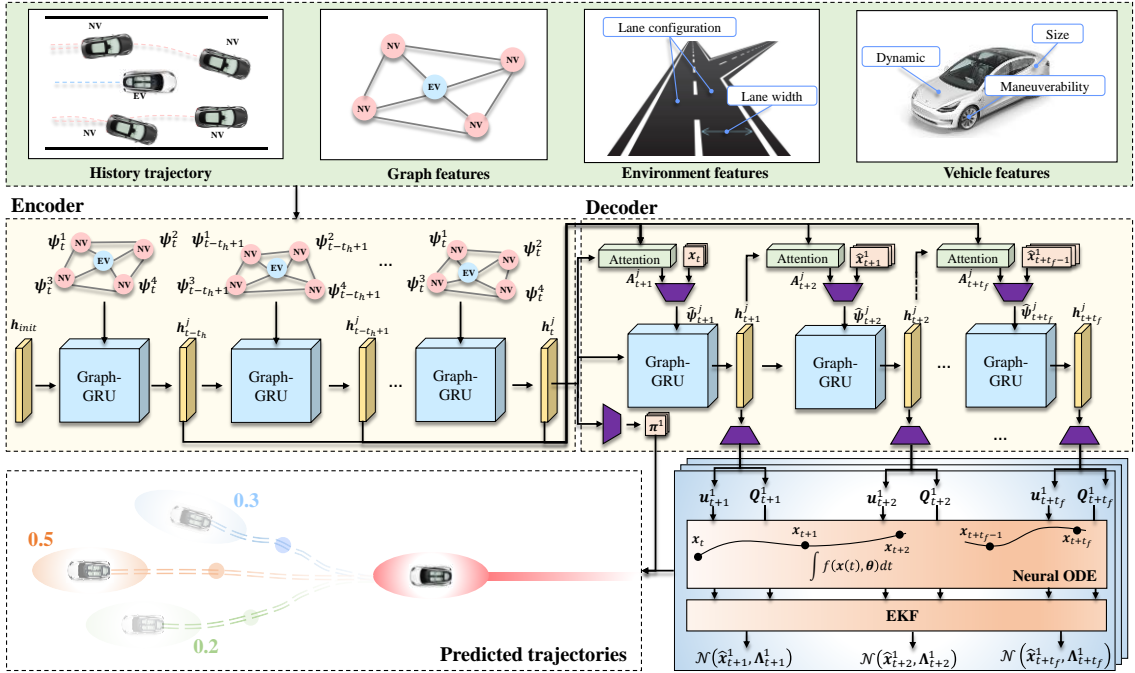

Fig. 5. Schematic illustration of the deep learning-based trajectory prediction architecture.

For each ego vehicle $i$ at time $t$, we construct a local interaction graph $\mathcal{G}_t^i = (V_t^i, E_t^i)$, where the node set is defined as $V_t^i = \{i\} \cup \mathcal{N}_i(t)$, denoting the set of neighboring agents present in the scene at time $t$, excluding the ego vehicle. The edge set $E_t^i$ captures the pairwise interactions among agents, which can be determined based on spatial proximity, relative velocity, or learned attention mechanisms. Accordingly, the historical observation for ego vehicle $i$ is defined as:

$$\mathcal{H} = \{(\mathcal{G}_t^i, \{\boldsymbol{\psi}_s^j | j \in \mathcal{N}_i\}) | s = t - t_h + 1, \ldots, t\} \tag{13}$$

where, $\boldsymbol{\psi}_s^j$ represents the feature vector of a neighboring agent $j \in \mathcal{N}_i$. This formulation captures both structural and motion information over the past $t_h$ time steps. The feature vector typically includes dynamic and semantic attributes such as position, velocity, acceleration, and the lateral offset from the lane centerline.

The process of encoding each agent's state and the interactions with its neighbors at each time step can be formalized as follows. First, the feature vectors of agent $j$ and its neighbor $w$ at time step $s$ are updated using the GNN module:

$$[\boldsymbol{\kappa}_{j,s}^r, \boldsymbol{\kappa}_{j,s}^z, \boldsymbol{\kappa}_{j,s}^h] = GNN_f(\boldsymbol{\psi}_s^j, \{\boldsymbol{\psi}_s^w\}_{j \neq w}) \tag{14}$$

This update captures both spatial and temporal dependencies by aggregating information from the agent's neighbors. Similarly, the hidden state is updated at each time step by considering the previous hidden states and the neighbors' states:

$$[\boldsymbol{\xi}_{j,s}^r, \boldsymbol{\xi}_{j,s}^z, \boldsymbol{\xi}_{j,s}^h] = GNN_h(\boldsymbol{h}_s^{j-1}, \{\boldsymbol{h}_s^{w-1}\}_{j \neq w}) \tag{15}$$

To update the hidden state of agent $j$ at time step $s$, we employ a gated mechanism, consisting of a reset gate $\boldsymbol{r}_s^j$ and an update gate $\boldsymbol{z}_s^j$. These gates help control the flow of information from previous states and current features. The reset gate $\boldsymbol{r}_s^j$ is computed as:

$$\boldsymbol{r}_s^j = \sigma\left(\boldsymbol{\kappa}_{j,s}^r + \boldsymbol{\xi}_{j,s}^r + \boldsymbol{b}_r\right) \tag{16}$$

where, $\sigma$ is the sigmoid function and $\boldsymbol{b}_r$ is the bias term. Similarly, the update gate $\boldsymbol{z}_s^j$ is computed as:

$$\boldsymbol{z}_s^j = \sigma\left(\boldsymbol{\kappa}_{j,s}^z + \boldsymbol{\xi}_{j,s}^z + \boldsymbol{b}_z\right) \tag{17}$$

where, $\boldsymbol{b}_z$ is also a bias term. This gate controls how much of the previous hidden state should be retained. The candidate hidden state $\widetilde{\boldsymbol{h}}_s$ is then calculated as:

$$\widetilde{\boldsymbol{h}}_s^j = \tanh(\boldsymbol{\kappa}_{j,s}^h + \boldsymbol{r}_s^j \odot \boldsymbol{\xi}_{j,s}^h + \boldsymbol{b}_h) \tag{18}$$

where, $\boldsymbol{r}_s^j$ acts as a gating mechanism to adjust the influence of past information on the current state. Finally, the current hidden state $\boldsymbol{h}_s^j$ is updated using the following equation:

$$\boldsymbol{h}_s^j = \left(1 - \boldsymbol{z}_s^j\right) \odot \widetilde{\boldsymbol{h}}_s^j + \boldsymbol{z}_s^j \odot \boldsymbol{h}_{s-1}^j \tag{19}$$

where $\boldsymbol{z}_s^j$ controls the proportion of the new hidden state $\widetilde{\boldsymbol{h}}_s^j$ and the previous hidden state $\boldsymbol{h}_{s-1}^j$ that are combined to form the final state.

The decoder aims to predict future trajectories by gradually generating the control input $\boldsymbol{u}_s$ at each time step $s$, which serves as the driving term for the motion model, while maintaining the graph structure unchanged. To enhance the influence of historical representations on the current prediction, an attention mechanism is introduced.

$$\boldsymbol{a}_{j,s}^{p} = \frac{exp(\boldsymbol{q}_{j,s}^{p})}{\sum_{s=t-t_h}^{t} exp(\boldsymbol{q}_{j,s}^{p})}, for\ s = t - t_h, \dots, t \tag{20}$$

where, $\boldsymbol{a}_{j,s}^{p}$ denotes the attention weight assigned by agent $j$ to the historical time step $s \in [t - t_h, t]$ when making predictions for time step $p$. It reflects the relative importance of past observation $s$ in contributing to the current prediction; $\boldsymbol{q}_{j,s}^{p}$ represents the compatibility score between the agent's prediction at time $p$ and its past information at time $s$, typically computed by an attention scoring function. The denominator performs a softmax normalization over all historical steps to ensure that the attention weights form a valid probability distribution, summing to 1 and ranging within [0,1]. The attention-weighted representation over historical time steps can be formulated as:

$$\boldsymbol{A}_i^v = \sum_{s=t-t_h}^{t} \boldsymbol{a}_{j,s}^{p} \cdot \boldsymbol{h}_s^j \tag{21}$$

During the prediction process, the uncertainty of the trajectory progressively accumulates over time. Given the smooth evolution of vehicle trajectories over short time horizons, the Extended Kalman Filter (EKF) provides reliable prediction performance and is thus employed to model this uncertainty propagation through its time update equations:

$$\hat{\boldsymbol{x}}_{p+1} = f(\hat{\boldsymbol{x}}_p, \boldsymbol{u}_p) \tag{22}$$

$$\boldsymbol{\Lambda}_{p+1} = \boldsymbol{F}_p \boldsymbol{\Lambda}_p \boldsymbol{F}_p^{\mathrm{T}} + \boldsymbol{G}_p \boldsymbol{Q}_p \boldsymbol{G}_p^{\mathrm{T}} \tag{23}$$

Here, $\boldsymbol{F}_p = \frac{\partial f}{\partial x}$ denotes the Jacobian matrix of the state transition function for the state variable, $\boldsymbol{Q}_p$ represents the process noise covariance matrix whose parameters are predicted by the model, and $G_k$ is the process noise input matrix, typically determined by the sampling period. The pair $(\hat{\boldsymbol{x}}_p, \boldsymbol{\Lambda}_p)$ represents the estimated state and its associated covariance. This evolution of the covariance matrix reflects the expansion of uncertainty in the state distribution over future time steps, thereby laying a foundation for downstream risk propagation modeling.

Based on this formulation, the model outputs at each prediction step $p$ are defined as:

$$\boldsymbol{y}_p = \left(\boldsymbol{\pi}^l, \{\hat{\boldsymbol{x}}_p^l, \boldsymbol{\Lambda}_p^l\}_{l=1}^{L}\right) \tag{24}$$

where, $\pi^l$ denotes the probability of the $l$-th modality. To enable the model to learn both accurate predictions and well-calibrated uncertainty estimates, a negative log-likelihood (NLL) loss is employed as the training objective:

$$\mathcal{L}_{NLL} = \sum_{t=1}^{t_f} -log\left(\sum_{l=1}^{L} \boldsymbol{\pi}^l \cdot \mathcal{N}_i\left(\boldsymbol{x}_p \middle| \hat{\boldsymbol{x}}_p^l, \boldsymbol{\Lambda}_p^l\right)\right) \tag{25}$$

### 4.2.3 Probabilistic risk prediction

Given that the output of the trajectory prediction model consists of discrete position sequences $\hat{\boldsymbol{x}}_p^l$, the average velocity is approximated using finite differences between the current observed position and the predicted future positions. Specifically, the estimated average velocity of the $l$-th trajectory mode at prediction

time step $p$ is computed as:

$$\hat{v}_{j,p}^l \approx \frac{\hat{x}_{j,p}^l - x_j(t)}{p \cdot \Delta t} \tag{26}$$

To determine the future positions and velocities of traffic participants $j$ for subsequent risk modeling, we define an interaction energy metric based on relative velocity. At each prediction time step $p$, the unit-mass interaction energy is formulated as:

Similarly, to account for the anisotropic propagation of risk along different velocity directions, an angular correction mechanism based on relative motion direction is introduced. Specifically, the angle between the ego vehicle's velocity and the predicted velocity is computed as:

$$\theta_{ij}^l(p) = \angle(v_i(t), \hat{v}_{j,p}^l) \tag{27}$$

At this point, for each trajectory mode $l$, the predicted future state $(\hat{v}_{j,p}^l, \hat{x}_{j,p}^l, \theta_{ij}^l(p))$ can be used as the input for risk modeling. Given that the multimodal trajectory prediction reflects the probabilistic distribution of possible future behaviors, the overall expected risk impact from participant $j$ at time $t + \tau$ can be obtained by weighting the risk field contributions of all modes:

$$\mathbb{E}[\tilde{F}_{ij}(p)] = \sum_{l=1}^{L} \pi^l \cdot \tilde{F}_{ij}^l(p) \tag{28}$$

Furthermore, considering parallel interactions with multiple participants, the total expected risk intensity for the ego vehicle at time $t + \tau$ is defined as:

$$\tilde{F}_i(p) = \sum_{j \in \mathcal{N}_i} I_{ij} \cdot \mathbb{E}[\tilde{F}_{ij}(p)] \tag{29}$$

To comprehensively assess the risk evolution over the entire prediction horizon, a time-weighted cumulative risk metric over the prediction period $t_f$ is defined as:

$$R_i^{total} = \sum_{p=1}^{t_f} \omega_r \cdot \tilde{F}_i(p) \tag{30}$$

where $\omega_r$ denotes the risk weight at each time step, which can be configured based on the growth of prediction uncertainty, the urgency of response, or task-oriented strategies.

## 5. Results and discussion

*5.1 Validation dataset*

To validate the proposed RiskNet framework, we conducted experiments on three real-world traffic datasets: highD, inD, and rounD. All were collected in German traffic environments under naturalistic conditions, covering a range of scenarios representative of the human-vehicle-infrastructure system and providing high-quality data for risk assessment in autonomous driving. The highD dataset includes 60 highway scenes and over 110,000 vehicle trajectories across various traffic states, free-flow, congestion, and lane changes. Captured via drones, it offers high-precision data on position, speed, acceleration, and lane-level behavior, making it ideal for highway risk analysis. As shown in Fig. 6, the inD dataset captures four urban intersections with over 5,000 vulnerable road users (VRUs) and 11,500 trajectories. It provides detailed motion and interaction data among vehicles, pedestrians, and cyclists, enabling urban risk and decision-making studies.

The rounD dataset focuses on roundabouts of varying sizes and flow conditions. It records full trajectories, speed, acceleration, and lane changes during entry, circulation, and exit, capturing complex interactions essential for modeling high-risk driving scenarios.

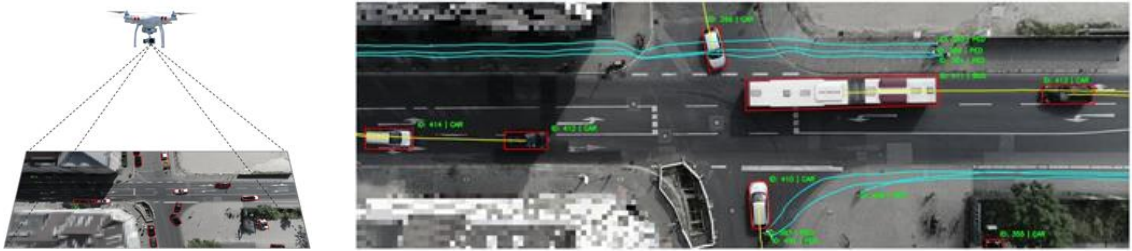

(a) Data collection process　　　　　　(b) Sample trajectories of traffic participants

Fig. 6. Overview of the inD dataset and trajectory examples

*5.2 Model validation*

5.2.1 Results analysis under deterministic scenarios

**Fig. 7** presents the quantified risk dynamics as the ego vehicle (ID: 679) performs a lane change in a highD highway scenario using the proposed interaction-field-based multisource risk modeling framework. The top panel shows spatial risk distribution, where warmer colors indicate higher risk. During the lane change, the model identifies multiple surrounding vehicles (e.g., IDs 674, 676, 686) also changing lanes or shifting laterally, increasing interaction intensity and creating overlapping high-risk zones. The main risk sources stem from trajectory convergence and speed differentials, which amplify collision potential. The middle plot shows the temporal risk evolution: initially stable, risk spikes during the mid-phase due to lateral interference, then declines as vehicles separate. The bottom panel displays vehicle trajectories in the X–Y plane, revealing spatial convergence. The model accurately captures both longitudinal and lateral interaction risks, achieving high directional sensitivity and timely recognition of lane-level risk accumulation during dynamic maneuvers.

**Fig. 8** illustrates the risk assessment results for typical ego vehicle maneuvers at intersections, including right turns, merging into main traffic flow, and left-turn yielding. The proposed interaction-field-based multisource risk modeling framework is used for quantitative evaluation. The top panel shows spatial risk distributions, with red regions indicating high-risk zones. The model successfully identifies risk sources from multiple directions, including the sudden appearance of a crossing pedestrian (ID: 138) and aggressive merging or acceleration behaviors by nearby vehicles (e.g., IDs 189, 164, 195). These interactions converge at the intersection, creating short-duration, high-intensity multi-directional conflicts and localized risk concentrations. The middle panel presents the temporal evolution of risk. Risk levels increase as the ego vehicle approaches the intersection or initiates turning maneuvers, peaking during interaction with cross traffic or pedestrians, then gradually decreasing as the vehicle completes its maneuver. The bottom panel shows the trajectory evolution of surrounding participants. Multiple vehicles perform distinct actions, such as merging, yielding, or turning—in close proximity, resulting in dynamic interaction conflicts and elevated collision risk.

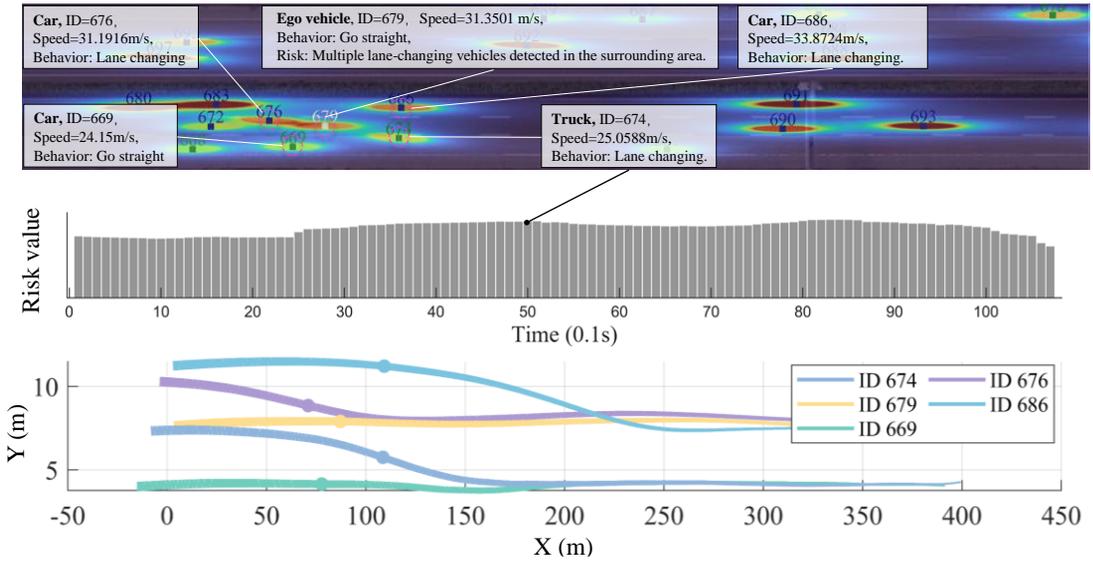

Fig. 7. Risk quantification during ego vehicle lane change (highD Dataset)

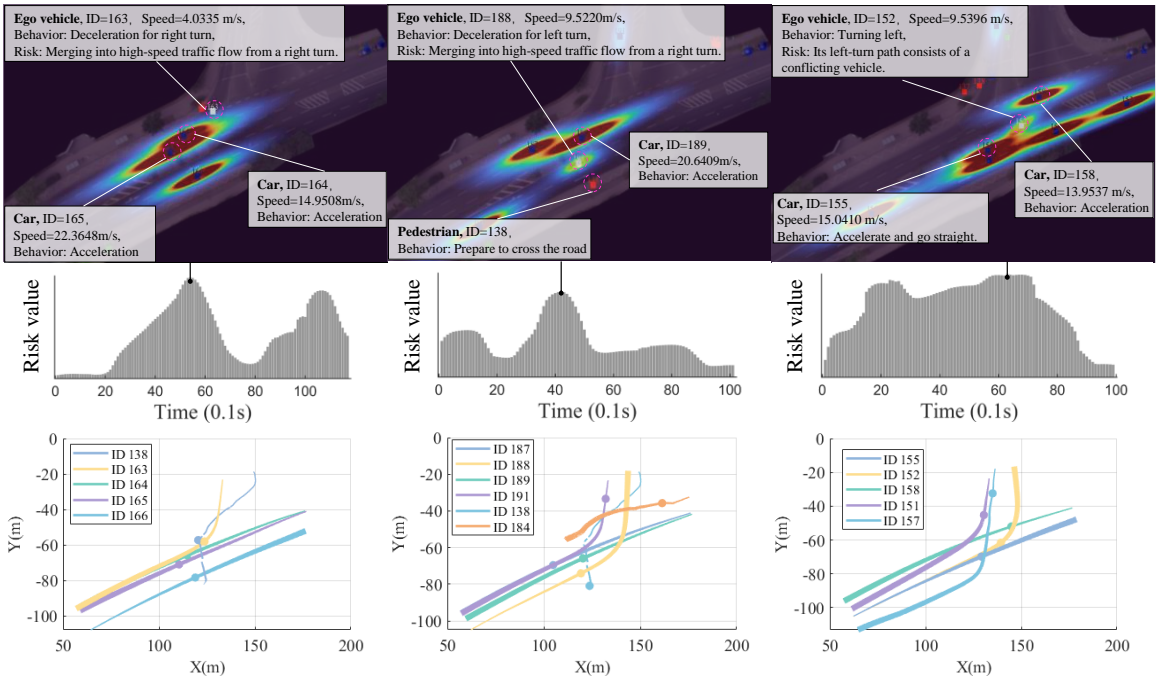

Fig. 8. Risk quantification during ego vehicle turning (inD Dataset)

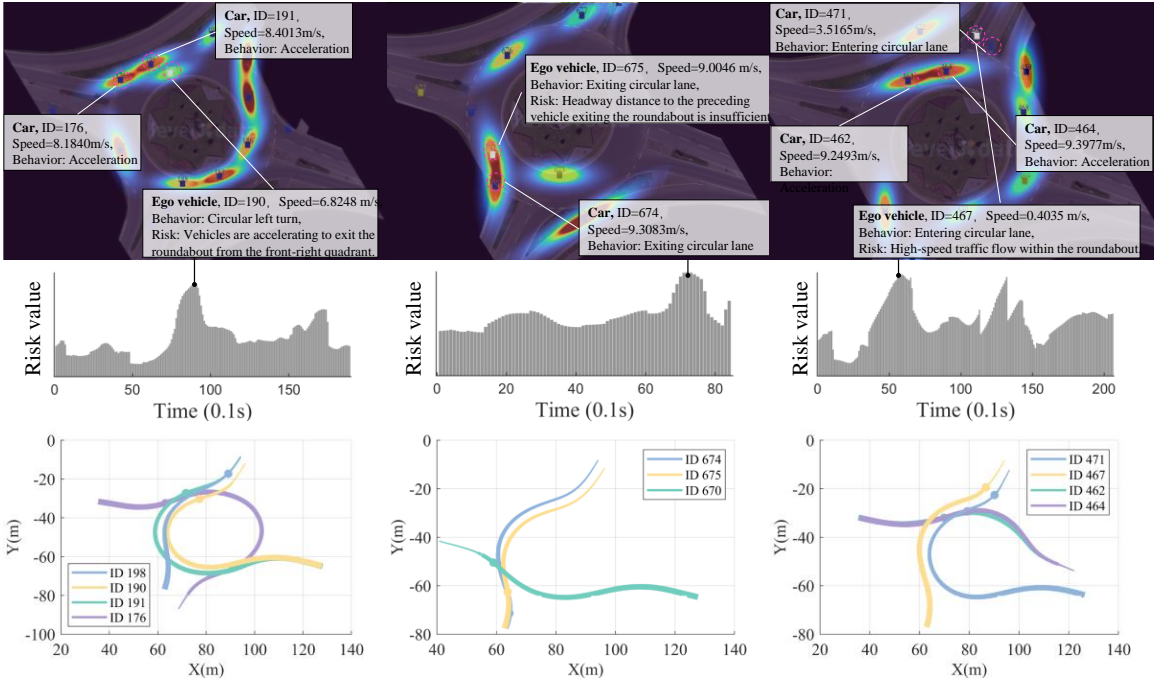

Fig. 9. Risk quantification in roundabout maneuvers (rounD Dataset)

**Fig. 9** shows risk assessment results for ego vehicle maneuvers, entering, circulating, and exiting a roundabout, using the proposed interaction-field-based deterministic risk model. The top panel displays spatial risk maps, with red zones indicating high-risk areas. The model captures multi-directional threats from merging vehicles (e.g., ID: 147), accelerating vehicles within the roundabout (e.g., IDs 130, 191), and conflicts at merging points. These interactions in constrained space create concentrated, direction-sensitive risks. The middle panel shows temporal risk evolution, with sharp increases during the entry and merging phases. The bottom panel illustrates vehicle trajectories, highlighting merging, yielding, and side-by-side movement. The model effectively combines directional, spatial, and kinematic cues to quantify surrounding vehicle influence, showing strong adaptability in complex scenarios.

### 5.2.2 Results analysis under uncertain scenarios

1) Multimodal trajectory prediction results

**Fig. 10** presents the multimodal trajectory prediction results from our GNN-based forecasting module across four representative driving scenarios. The gray line indicates historical positions, the black "×" the last observed point, the blue dashed line the ground truth future trajectory, and the colored dots denote predicted trajectories across modalities $l$, with transparency indicating their associated probabilities $\pi^l$.

The top-left shows a low-uncertainty, straight-driving case, where the prediction aligns closely with the true trajectory and exhibits a concentrated distribution. The top-right case involves a slight deceleration or lane curvature, where the model identifies two dominant trajectory hypotheses. The primary mode remains closely aligned with the ground truth, highlighting the model's ability to distinguish subtle behavior variations. The bottom-left case illustrates a sharp turning maneuver with pronounced curvature. Despite high trajectory nonlinearity, the model successfully captures the turning behavior, and the top mode closely follows the true trajectory, demonstrating robustness in modeling non-linear motion. The bottom-right shows moderate acceleration with mild steering, yielding compact and accurate predictions with dominant confidence. Overall, the proposed GNN-based predictor exhibits strong multimodal forecasting capability and effective uncertainty modeling. It reliably anticipates diverse agent behaviors under uncertain conditions, supporting downstream risk assessment and decision-making modules with high-fidelity future motion hypotheses.

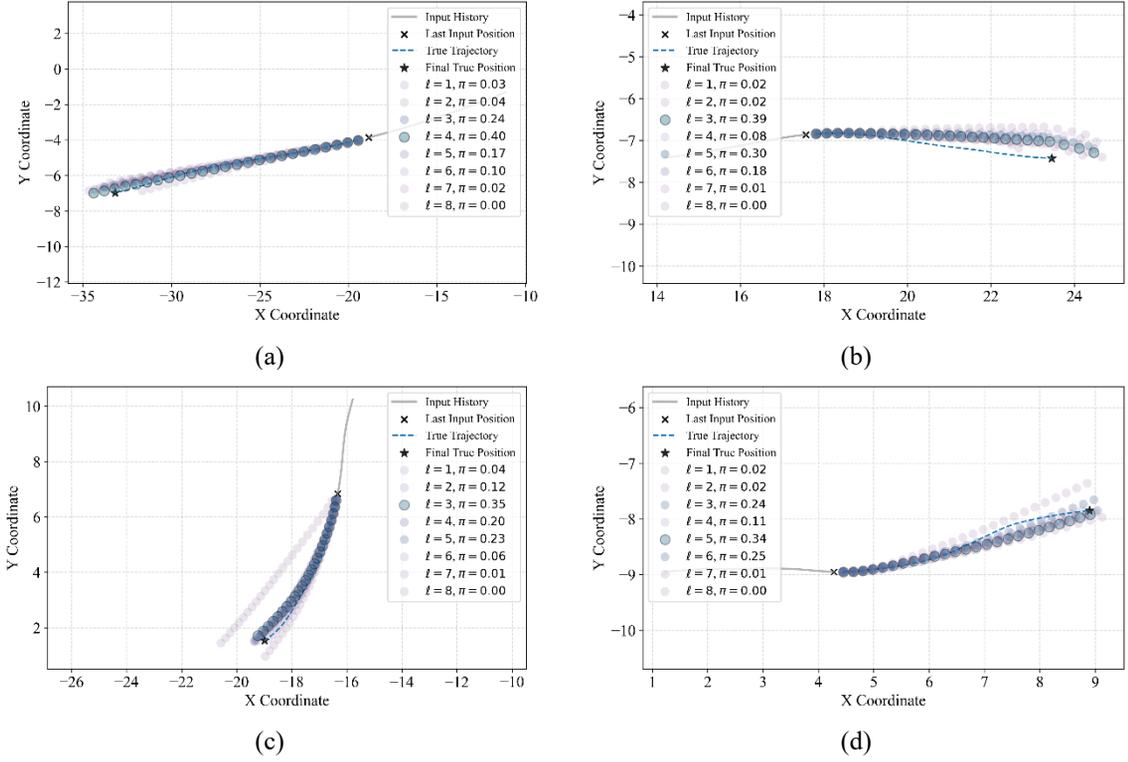

Fig. 10. Multimodal trajectory prediction results under uncertain scenarios.

Table 1 presents the quantitative results corresponding to the multimodal predictions shown in Fig. 10. The model achieves strong prediction accuracy, with an Average Displacement Error (ADE) of 0.44 m and a Final Displacement Error (FDE) of 1.35 m, indicating effective trajectory fitting. The Average Probabilistic Displacement Error (APDE) of 0.26 m further confirms the representativeness of the most likely predicted mode. In terms of uncertainty modeling, the Average Negative Log-Likelihood (ANLL) of -1.22 and the First-Mode Negative Log-Likelihood (FNLL) of 2.38 indicate reliable and coherent probabilistic outputs. Overall, the proposed GNN-based framework demonstrates strong multimodal prediction and uncertainty reasoning capabilities.

**Table 1** GNN-based trajectory prediction performance

| Metric | ADE (m) | FDE (m) | APDE (m) | ANLL (m) | FNLL (m) |
|---|---|---|---|---|---|
| Value | 0.44 | 1.35 | 0.26 | -1.22 | 2.38 |

2) Probabilistic risk map

**Fig. 11** presents a comparative analysis between the proposed probabilistic risk maps and traditional deterministic risk representations under two representative dense interaction traffic scenarios. Subfigures (a-1) and (a-2) illustrate a scenario where a high-speed vehicle abruptly approaches from a side road at an intersection. The probabilistic risk map (a-1), informed by trajectory uncertainty, captures a broader and more continuous risk region that reflects the threat posed by multiple potential paths of the high-speed vehicle. In contrast, the deterministic map (a-2) concentrates risk narrowly along a single predicted trajectory, potentially underestimating the overall hazard. Subfigures (b-1) and (b-2) correspond to a rear vehicle accelerating to merge ahead of the ego vehicle. In (b-1), the probabilistic map reveals multiple high-risk hotspots around the interaction region, showcasing the model's capacity to account for behavioral uncertainty and multimodal disturbance. Meanwhile, the deterministic map (b-2) focuses only on the specific predicted path, lacking responsiveness to alternative maneuvers. These results demonstrate that probabilistic risk maps better capture uncertain and evolving threats, supporting safer decision-making in complex environments.

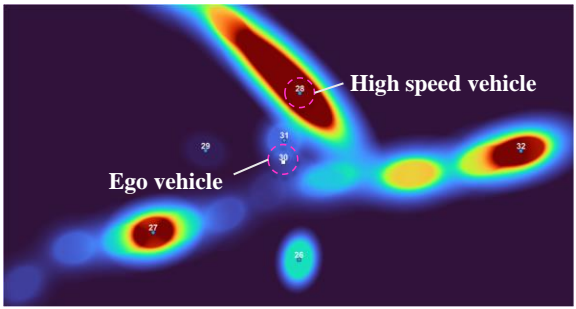

(a-1) Probabilistic risk map: interaction conflict

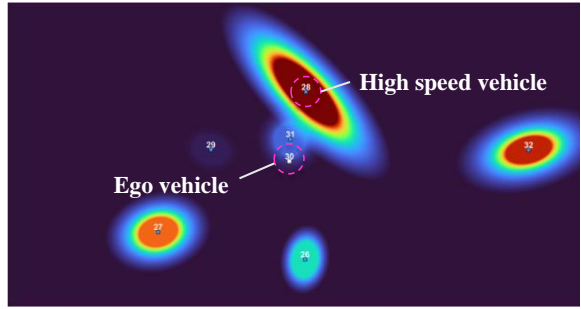

(a-2) Deterministic risk map: interaction conflict

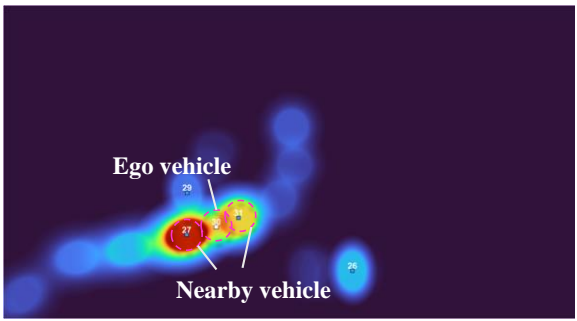

(b-1) Probabilistic risk map: rear vehicle conflict

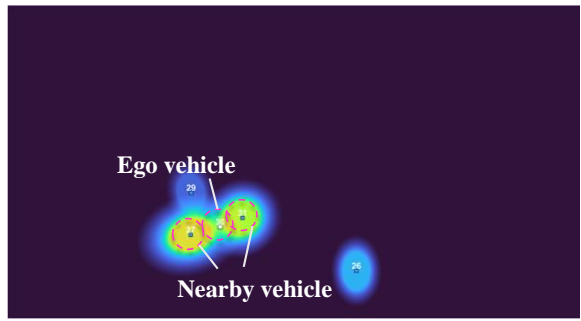

(b-2) Deterministic risk map: rear vehicle conflict

Fig. 11. Probabilistic vs. deterministic risk map comparison in complex interaction scenarios

5.2.3 Comparative results anaylsis

Table 2 summarizes three representative vehicle–vehicle interaction scenarios, covering high-risk behaviors such as lane changing, cut-ins, and lateral merges. These scenarios reflect key challenges faced by autonomous driving systems in long-tail situations. Each involves multi-agent dynamics, with clearly defined risk sources and multi-directional conflict patterns. Such complex real-world interactions provide a meaningful basis for evaluating the proposed method's capability in recognizing risks and maintaining robustness across diverse interaction types.

**Table 2** Overview of vehicle interaction scenarios

| ID | Ego Type | Ego Intention | Vehicle Count | Other Types | Main Risk Source | Key Frame |
|---|---|---|---|---|---|---|
| Scenario 1 | Car | Lane change | 3 | Car | Lane change blocked by vehicles in front and rear gaps. | 100-200 |
| Scenario 2 | Truck | Lane keeping | 2 | Car\Truck | Side vehicle accelerates and merges, causing lateral conflict. | 100-300 |
| Scenario 3 | Car | Lane keeping | 2 | Car | Rear vehicle overtakes and cuts in, causing longitudinal pressure. | 10-140 |

**Fig. 12** illustrates the risk recognition process as the ego vehicle (in red) attempts a right lane change. The primary risk arises from three surrounding vehicles: one slow-moving vehicle in the target lane and two fast-approaching vehicles in the ego lane, forming a constraining front-rear interaction. TTC and THW, which focus solely on longitudinal metrics, are highly sensitive to the front and rear vehicles before the lane change, continuously flagging high risk. However, they fail to detect the slow vehicle in the target lane during the lane change itself. RSS briefly detects risk early in the sequence but underestimates it in the later stages. NC Field does not account for rearward threats and responds with significant delay. In contrast, our proposed method accurately identifies the compounded lane-change risk throughout the interaction and maintains a consistent high-risk output, demonstrating stronger robustness and directional sensitivity.

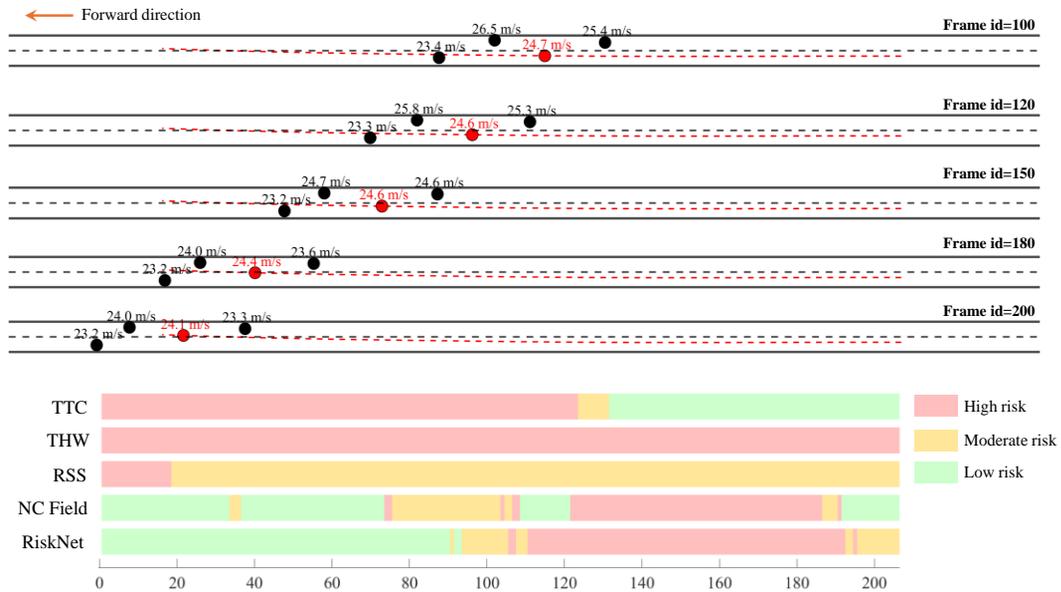

Fig. 12 Scenario 1: risk assessment during ego vehicle lane change

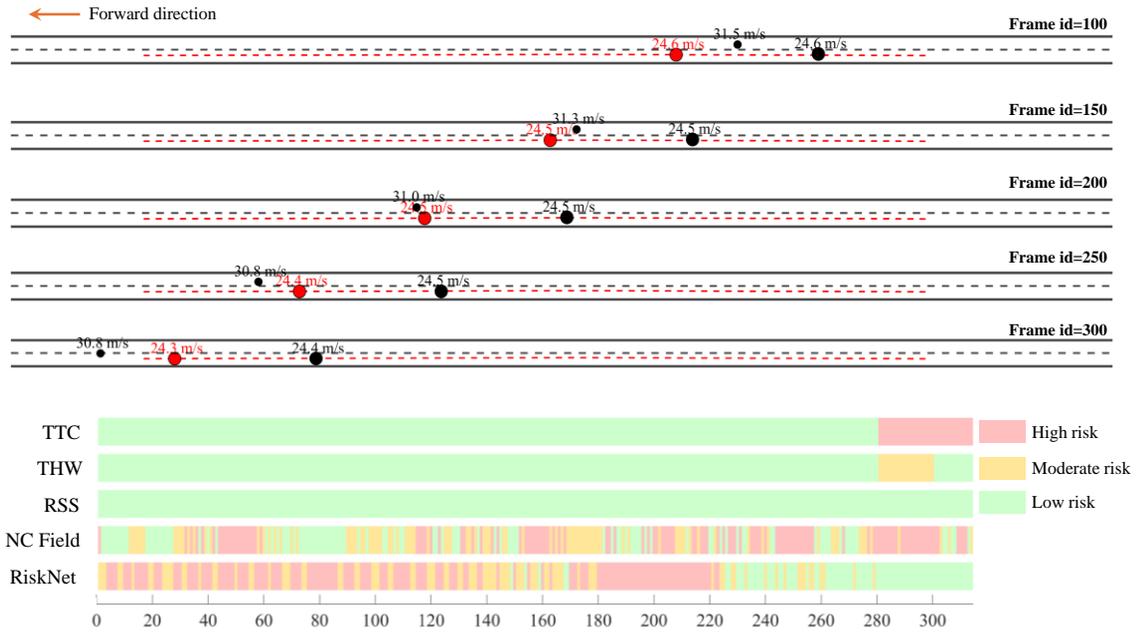

Fig. 13 Scenario 2: risk assessment for right-side cut-in maneuver

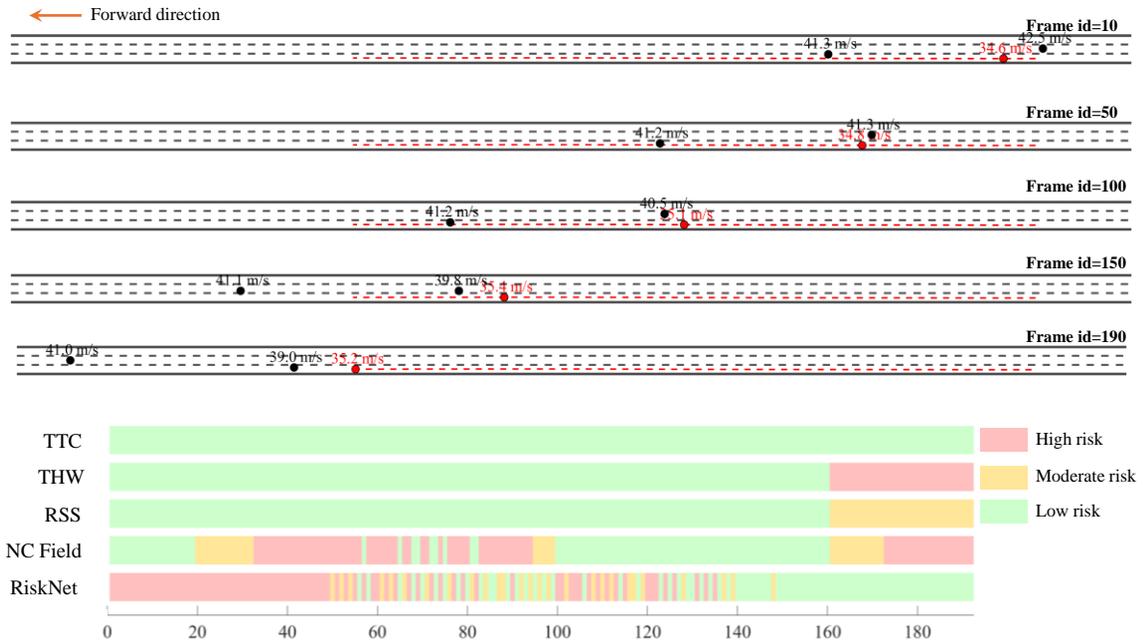

Fig. 14 Scenario 3: risk assessment for high-speed rear overtake and cut-in

As shown in **Fig. 13**, the ego vehicle (red), a truck, maintains a steady cruising speed in its lane, while a vehicle on the right (black) suddenly accelerates and merges laterally, introducing a high-risk disturbance. This risk primarily arises from the abrupt lateral intrusion coupled with a significant longitudinal velocity difference. Conventional indicators such as TTC and THW fail to capture this interaction, as they rely solely on front-rear proximity and neglect lateral dynamics. Moreover, these indicators do not account for critical physical characteristics such as vehicle shape, mass, and kinematic properties, which are essential for realistic risk estimation in heterogeneous traffic. RSS offers a brief response near the merging point but lacks temporal continuity and spatial adaptability. While NC Field demonstrates partial lateral sensitivity, it exhibits clear limitations in perceiving risks emerging from the rear and lateral-rear zones of the ego vehicle. Specifically, it underperforms in scenarios where trailing vehicles accelerate and overtake from behind, missing early indicators of potential conflict. In contrast, our proposed method identifies the cut-in risk before the merging vehicle physically enters the ego lane and maintains consistent high-risk prediction throughout the interaction. This reflects superior responsiveness, stronger directional awareness, and enhanced sensitivity to dynamic multi-agent interactions.

In **Fig. 14**, the ego vehicle (red) maintains a steady lane position while a faster rear vehicle (black) approaches and suddenly cuts in from the side, creating a compound risk involving both longitudinal pressure and lateral intrusion. This behavior is abrupt and deviates from normal yielding patterns. TTC and THW fail to detect the early-stage risk, reacting only briefly after the vehicle enters the ego lane. RSS provides a moderate warning upon cut-in but lacks sensitivity during the approach phase. NC Field, lacking rearward modeling, responds too late. In contrast, our method anticipates the risk before the overtake is completed, leveraging interaction fields and directional corrections. It captures intent shifts based on speed and spacing, outputting continuous high-risk estimates aligned with real-world collision dynamics.

## 6.  Conclusions

This paper addresses the critical challenge of risk quantification for AVs operating in long-tail scenarios—low-frequency yet high-severity conditions characterized by behavioral uncertainty and complex multi-agent interactions. To tackle this, we introduce RiskNet, an interaction-aware risk forecasting framework that integrates physics-based deterministic modeling with deep learning-based probabilistic trajectory prediction. By combining interaction fields with multimodal behavior forecasting, RiskNet captures the coupled influence of infrastructure, vehicle dynamics, and behavioral uncertainty, enabling interpretable and scalable risk estimation in complex traffic environments. Comprehensive evaluations on the highD, inD, and rounD datasets, covering highways, intersections, and roundabouts, show that RiskNet consistently outperforms classical metrics such as TTC, THW, RSS, and NC Field in accuracy, spatial sensitivity, and efficiency. The framework generalizes robustly across high-interaction scenarios and road configurations, maintaining accuracy in detecting emergent conflicts and spatially distributed risk under long-tail conditions.

Future work will focus on enhancing adaptability and closed-loop integration. First, we will incorporate sensor-level uncertainty, such as occlusion, misdetection, and communication delay, to improve robustness under multi-source disturbances. Second, we aim to develop risk-aware planners that leverage the predicted risk field to proactively avoid high-risk zones, enabling safer and more adaptive behavior (Liu et al., n.d.; Q. Liu et al., 2024). This direction opens the possibility of a tightly coupled pipeline from risk perception to planning and control, advancing toward risk-driven autonomous behavior in multi-agent environments.